\documentclass{article}

\usepackage{arxiv}

\usepackage[utf8]{inputenc} 
\usepackage[T1]{fontenc}    
\usepackage{hyperref}       
\usepackage{url}            
\usepackage{booktabs}       
\usepackage{amsfonts}       
\usepackage{nicefrac}       
\usepackage{microtype}      
\usepackage{lipsum}
\usepackage{tikz} 
\usepackage{amsmath}
\usepackage{float}
\usepackage{stfloats}
\usepackage{threeparttable}
\usepackage{amsmath}
\usepackage{enumitem}
\usepackage{soul}
\usepackage{caption}
\usepackage{subcaption}
\usepackage{float}
\usepackage{booktabs}
\usepackage{footmisc}
\usepackage{graphicx}

\setlist[description]{leftmargin=0 pt}

\title{Guess what's on my screen? \\ Clustering Smartphone Screenshots with Active Learning}

\author{
  Agnese Chiatti \\
  Information Sciences and Tenchology \\
  The Pennsylvania State University\\
  \texttt{azc76@psu.edu} \\
   \And
 Dolzodmaa Davaasuren \\
 Industrial and Manufacturing Engineering \\
 The Pennsylvania State University \\
  \texttt{dud240@psu.edu} \\
  \And
 Nilam Ram \\
 Human Development and Family Studies \\
 The Pennsylvania State University \\
  \texttt{nur5@psu.edu} \\
  \And
 Prasenjit Mitra \\
 Information Sciences and Tenchology \\
 The Pennsylvania State University\\
  \texttt{pmitra@ist.psu.edu} 
  \And
 Byron Reeves \\
  Department of Communication \\
 Stanford University\\
  \texttt{reeves@stanford.edu}
  \And
 Thomas Robinson \\
 Departments of Pediatrics and Medicine \\
 Stanford University\\
  \texttt{tom.robinson@stanford.edu}
}

\begin{document}
\maketitle

\begin{abstract}
A significant proportion of individuals' daily activities is experienced through digital devices.  Smartphones in particular have become one of the preferred interfaces for content consumption and social interaction. Identifying the content embedded in frequently-captured smartphone screenshots is thus a crucial prerequisite to studies of media behavior and health intervention planning that analyze activity interplay and content switching over time. \\
Screenshot images can depict heterogeneous contents and applications, making the \textit{a priori }definition of adequate taxonomies a cumbersome task, even for humans. Privacy protection of the sensitive data captured on screens means the costs associated with manual annotation are large, as the effort cannot be crowd-sourced. Thus, there is need to examine utility of unsupervised and semi-supervised methods for digital screenshot classification. This work introduces the implications of applying clustering on large screenshot sets when only a limited amount of labels is available.\\
In this paper we develop a framework for combining K-Means clustering with Active Learning for efficient leveraging of labeled and unlabeled samples, with the goal of discovering latent classes and describing a large collection of screenshot data. We tested whether SVM-embedded or XGBoost-embedded solutions for class probability propagation provide for more well-formed cluster configurations. Visual and textual vector representations of the screenshot images are derived and combined to assess the relative contribution of multimodal features to the overall performance.
\end{abstract}


\section{Introduction}
\begin{figure*}[t!]
    \centering
    \begin{subfigure}[t]{0.2\linewidth}
        \centering
        \includegraphics[width=\linewidth]{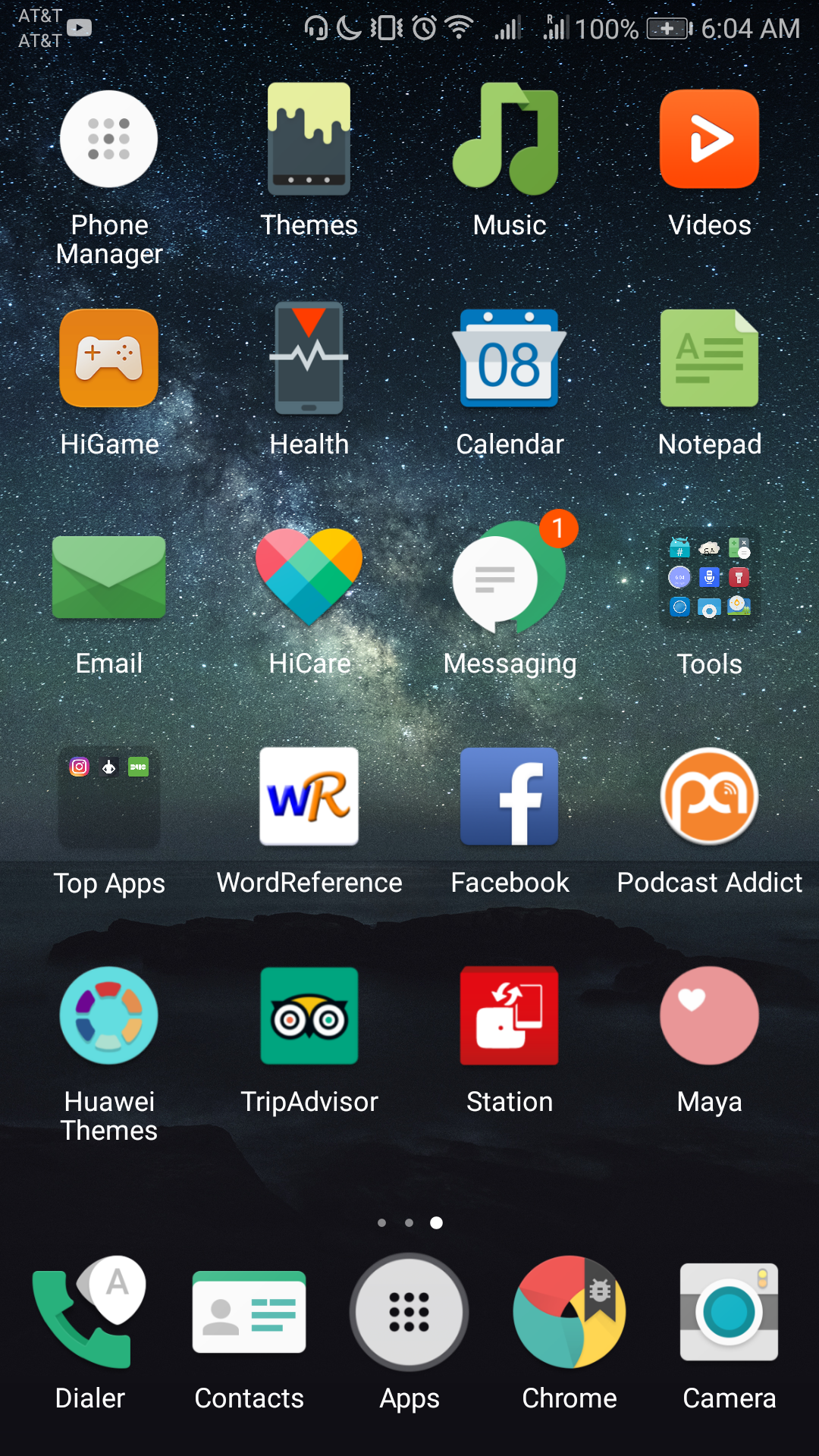}
        \caption{\label{fig:home}}
    \end{subfigure}%
    \hspace{1cm}
    \begin{subfigure}[t]{0.2\linewidth}
        \centering
        \includegraphics[width=\linewidth]{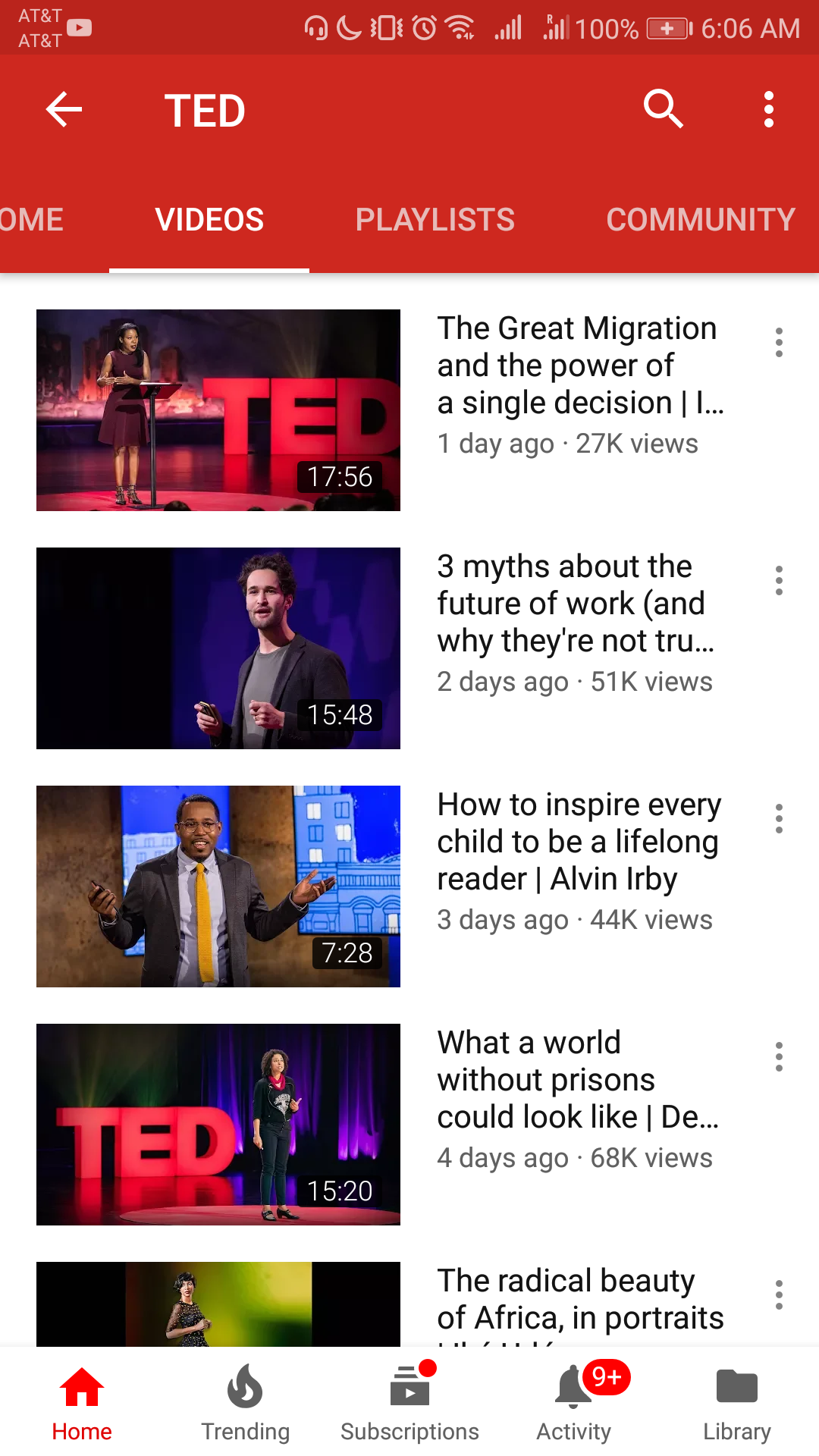}
        \caption{\label{fig:yt}}
    \end{subfigure}
        ~
        \hspace{1cm}
       \begin{subfigure}[t]{0.2\linewidth}
        \centering
        \includegraphics[width=\linewidth]{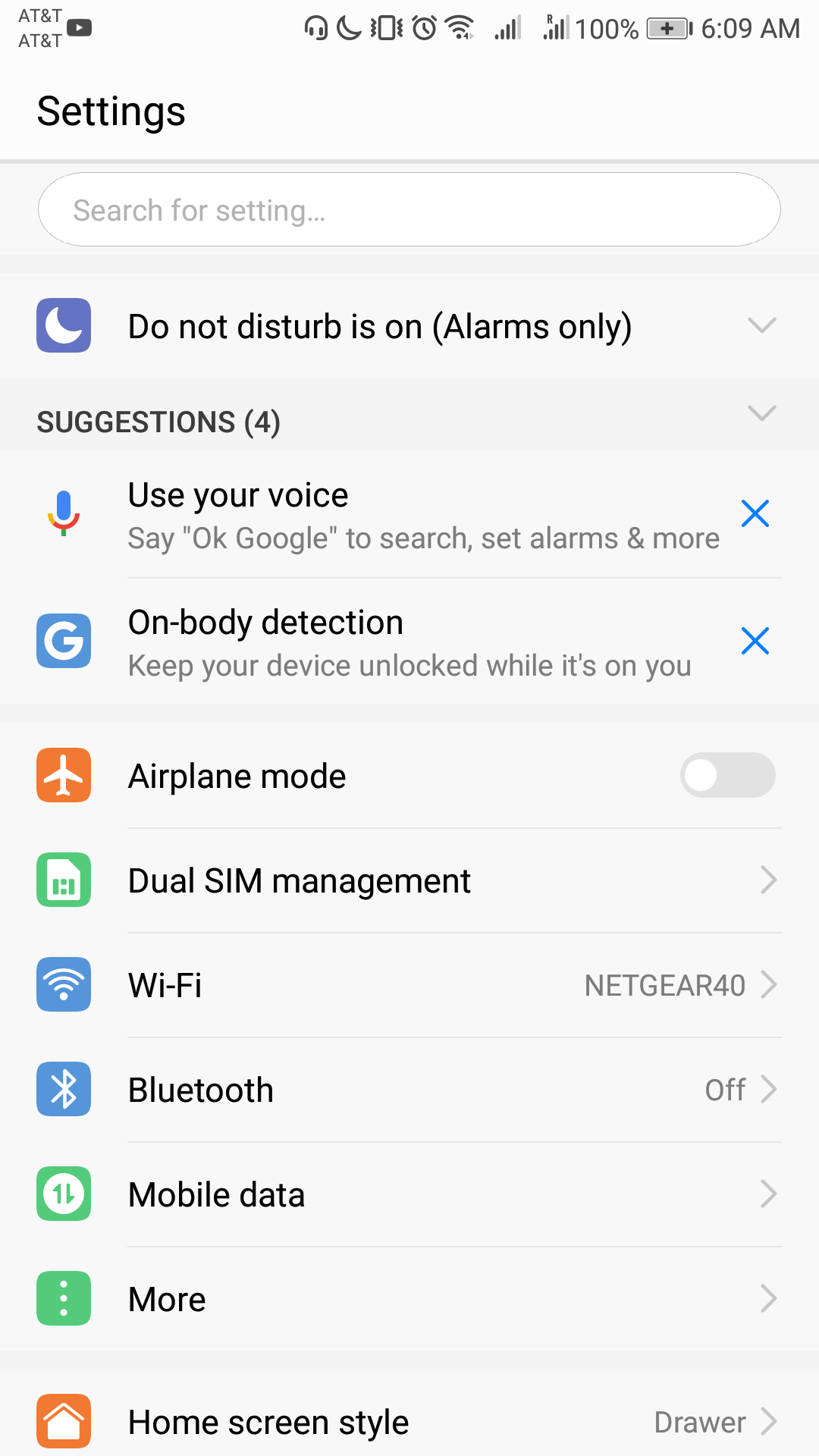}
        \caption{\label{fig:settings}}
    
    \end{subfigure}
    \caption{\label{fig:categories} Sample screenshots which would be labeled as: (a) Home screen (b) Youtube, and (c) Settings respectively.}
\end{figure*}

Daily experiences are increasingly experienced through digital devices and smartphones in particular have become a predominant site for consuming, sharing and searching 
for media contents. Thus, a number of studies of human behavior and "just-in-time" intervention planning can stem from intensive and longitudinal collection of smartphone screenshots,
%
%
which provide observations of individuals' second-by-second interactions with a wide variety of content.  
To successfully represent the fast-paced interplay of activities and contents constituting these life threads - with switches occurring as quickly as every 19 seconds \cite{Yeykelis2014}- behavioral researchers
%
%
need a taxonomy of screen content categories.  Paradigmatic use cases include, but are not limited to: early disease detection and prevention planning; HCI models of task switching and its implications for attention and memory; ethnographic assessments on the effects of marketing strategies and profiling on low-income populations; studies of political attitudes and voting expressions across social media and news media, and so forth. \\
%
%
Classification, however, can be cumbersome especially for screenshots,
that include nested data streams (images, text), presented over diverse templates.  For instance, if Figures \ref{fig:home} and \ref{fig:yt} are compared, one can see how the proportion of text over icons and graphical content is different. Plus, in Figure \ref{fig:yt} specifically, more complex frames (i.e., video previews) are alternated with textual contents and titles.  
%
%
As such, smartphone screenshots form a unique, yet rather unexplored, data typology, providing opportunities to test state-of-the-art methods for data clustering and classification on an unusual collection.   \\
%
%
Further, the confidential nature of screenshots harvested from individuals' daily phone use prompts searching for alternative options to crowd-sourcing for content categorization. 
%
%
Hence, there is much incentive to test state-of-the-art methods for Image Classification to automate this routine. Additionally, pre-defining adequate taxonomies capturing the full spectrum of relevant activities and applications is a challenging task. Screen behavior is 
far-reaching, stretching far beyond domain-specific research questions. 
%
%
%
%
For instance, annotating all sites for social networking as "Social Media" creates a rather amorphous category in a scenario where more social media platforms are being concurrently observed, or, as in this case, the evolution of information threads is observed over time. While, in fact, practitioners investigating the dissemination of media trends might be more interested in the social aspect of content sharing, researchers studying the expression of sentiment and voting attitudes across social feeds might find  "News" to be a more adequate category of interest for that same content.  \\
More suitable alternative taxonomies include tags that indicate the specific application being used (e.g., Facebook, Instagram, Twitter) or the typology of action being performed, when the specific application cannot be inferred (e.g., when watching videos in full screen).  As a result, to comprise for different activities and modalities inherent to the same "social media" app, we opted for application-level tags in such cases. In the longer run, more abstract categories could be then derived from the lower-level, application-based ones, based on the specific research hypotheses being tested.\\
%
%
Recognizing potential patterns from an heterogeneous interchange of applications (and contents, within the same app), as well as the requirement to leverage classification accuracy with a limited annotation budget, creates a need for an automated and high-level representation for screenshots that will generalize across different bounding conditions and efficiently integrate the multi-modal data sources embedded in those screenshots (i.e., the image itself and related textual contents).\\
The main contributions of this paper are as follows.
\begin{enumerate}[label=(\roman*)]
\item{We propose a framework that combines clustering with Active Learning and apply it on a novel data collection, i.e., digital screenshots captured from smartphone devices.}
\item{We introduce a joint representation of visual and textual features to assess the effects on the resulting cluster validity.}
\item{We discuss lessons learned inspiring future directions of improvement for this challenging classification task.}
\end{enumerate}

\section{Related work}
\subsection{Image Classification}
In the analyzed scenario, both the availability of a large collection of unlabeled screenshots and the costs associated with manual tagging, favor the choice of unsupervised and semi-supervised classification methods.  One first step toward classification is then discovering underlying patterns to group screenshots based on similarity between their most characteristic features and descriptors. 
Clustering techniques have been already proven useful in the related literature, when applied on unsupervised feature learning and classification tasks from unlabeled image collections. 
These methods have been applied independently \cite{hu2015unsupervised} or in combination with Deep Neural Networks \cite{yang2016joint}, to improve clustering through backward representation learning. Much effort has been devoted to the extraction of visual features as suitable descriptors for classification purposes \cite{zhang2017leaf} .\\ 
Further, K-Means clustering has been successfully applied to classify medical images \cite{antony2015new}, remotely-sensed images \cite{banerjee2015new} or, more broadly, natural scenes. As indicated in the literature, the random centroid initialization step embedded in the classic K-Means algorithm can affect the quality of results. However, this performance threat can be counteracted through exploiting more refined seeding strategies, such as K-Means++ \cite{arthur2007k}. \\
%
%
Compared to the reviewed image type, screenshots enclose heterogeneous contents whose nature is hard to classify. Nonetheless, we have found it to be a rather unexplored domain. The work conducted in \cite{Haskell2015CNNSystems}, is the most similar found to our case. The authors, Sampat et al., used a 3-layered convolutional network to classify a dataset of laptop screenshots over 14 classes. The architecture was pre-trained on the Places205CNN dataset from MIT, with the weights then transferred over to classify the target laptop screenshots. 
They concluded that the architecture was not effective for classification.
However the implications of training directly on screenshots (i.e., instead of using hyperparameters learned from more general image collections) were not addressed. 
Furthermore, our target images are still relatively different from the aforementioned set, as dealing with smartphone screenshots allows to skip the identification of the main active window on the screen, reducing the risk of error propagation from the latter step to the feature extraction phase.  

\subsection{Text Classification}
Distributed representation of words \cite{mikolov2013distributed} has been widely adopted for textual corpora vector transformations, both in supervised and in unsupervised scenarios. 
Recent work \cite{lilleberg2015support} has illustrated how combining word2vec representations with more traditional heuristics, i.e., namely TF-IDF, can aid linear separability in classification. Moreover, GloVe \cite{pennington2014glove},  a semantic vector space model based on term co-occurrence counts, has been shown to outperform the word2vec probabilistic approach on specific NLP sub-tasks, e.g. word analogy.\\
When dealing with images, the classification problem is usually tackled from a Computer Vision perspective, rather than relying on extracted textual contents. In fact, text extraction is error-prone, and can introduce more noise in the data analysis. On the other hand, images that are not visually similar, could
actually have similar contents. Furthermore, smartphone screenshot images are, on average, much more textually dense than  the images included in popular image collections \cite{russakovsky2015imagenet}. 
The evidence of segmentation defects on OCR-extracted text from digital screenshots \cite{chiatti2018sac}, 
%
%
%
%
support the use of text as an auxiliary component that can be used in classification. 

\subsection{Multi-modal Learning}
Joint-learning of image-text embeddings \cite{wang2016learning} has been applied to image datasets associated with transcribed captions
to improve entity localization.
Recent work has shown how multi-modality can lead to better knowledge capture, especially when extrapolating low-level concepts from the represented entities \cite{Both:2017:CKT:3148011.3148026}. In this context, text was pre-processed by applying known techniques in Text Classification, e.g. word2vec \cite{mikolov2013distributed} or GloVe \cite{pennington2014glove}, to be more uniformly merged with the image vectors \cite{bolanos2017vibiknet}. In 
other cases, visual patterns and textual features have also been learned jointly for automated event extraction \cite{zhang2017improving}.\\ 
However, to our knowledge, none of the reviewed multi-modal approaches have been tested on digital screenshots. In the related literature, these methods are typically applied on image caption learning \cite{mao2017deepart}, question answering tasks \cite{bolanos2017vibiknet}, or sematic triple representation learning \cite{nian2017multi}. Instead, we include the full textual data as a feature, and measure the relative impact of multi-modal feature sources on the resulting cluster validity. 

\subsection{Active Learning}
Active Learning methods have been thoroughly explored, to compensate for the lack of human annotation budget or, more broadly, maximize a task-oriented utility function. While many methods simply concentrated on single-valued definitions of model uncertainty, often combined with committee-based disagreement assessments, margin-based methods have provided a more refined alternative for image classification tasks \cite{tuia2011survey}. Classic definitions of margin with respect to SVM decision boundaries have also been adapted to the clustering case, to select the minimum margin for points that are most equidistant from their two closest subspaces \cite{lipor2015margin}. Further, the introduction of batch methods exploiting dynamic programming algorithms has facilitated scale towards larger data sets. To our knowledge, all of the surveyed methods have yet to be applied to digital screenshot images. \\
Inspired by different contributions, summarized in \cite{demir2011batch} and \cite{lipor2015margin}, and in the attempt to apply the lessons learned on high-resolution remote sensed images, we customized an Active Learning implementation that combines batch selection of informative points with diversity requirements\footnote{\url{https://github.com/google/active-learning}}, to suit the framework presented in this paper. 

\section{Semi-supervised Clustering with Active Learning}
\begin{figure*}[htbp]
\centering
\caption{\label{fig:framework} Proposed framework combining a joint and dimensionally-reduced vectorized representation for smartphone screenshots and their embedded textual contents. K-Means clustering is applied in combination with informative and diverse active learning to query a human oracle and predict probabilities for the unlabeled data points. }
\includegraphics[scale=0.3]{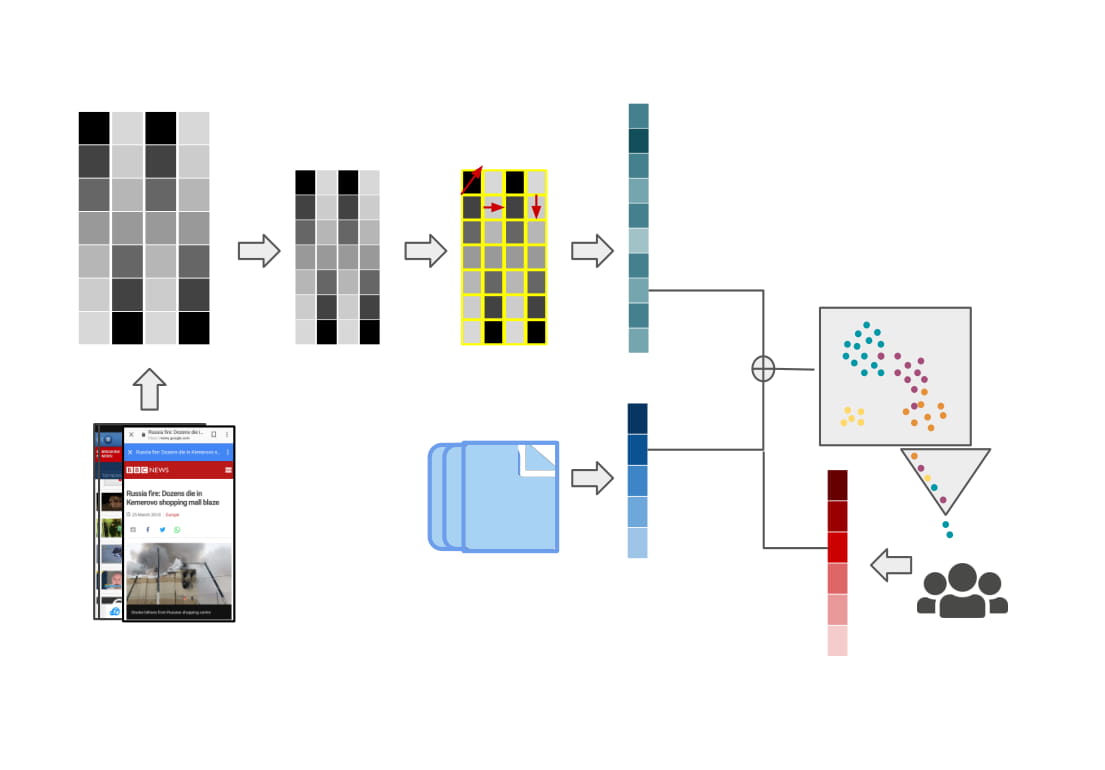}
\end{figure*}

The proposed workflow is illustrated in Figure \ref{fig:framework} and combines K-Means semi-supervised clustering with informative and diverse batch selection from a pool of unlabeled samples. The label information gathered by querying a human oracle is then leveraged with weaker probability estimates produced by a supervised classifier, for all the remaining points in the unlabeled set. 

\subsection{K-Means Clustering}
Historically introduced for signal processing, K-means is a partitional clustering method attempting to identify as many group centroids as specified by users through a dedicated parameter (K). Different setups can be chosen to initialize the centroids, e.g. K-Means++ \cite{arthur2007k}. Points are then clustered as a result of being assigned to their closest centroid. Centroids are updated based on the formed clusters, with assignments and updates repeated until convergence of the cluster configuration.\\ 
The framework experimented in this paper first applies unsupervised K-Means to select an initial batch of points through Active Learning. In subsequent iterations, labeled and unlabeled features are integrated into the vector representations so that clustering proceeds in a semi-supervised manner. 

\subsection{Informative and Diverse Active Learning} 
Active Learning methods can aid the selection of points to be labeled, and thus minimize the annotation effort. Two criteria framing the optimality of selected batches, with respect to some utility value (or labeling cost), are informativeness and diversity. While informativeness quantifies the ability to reduce the uncertainty on the underlying clustering model, another important contributing factor is ensuring disparity across the selected instances in order to mitigate potential redundancies. \\
The experimented framework adapted a released AL routine and combined it with K-Means clustering, to take both factors into account. First, informativeness is measured in a margin-based fashion. For each point $p$ in the input (set of size $N$), the related margin ($M$) is defined as:
\begin{equation}
M = d_i - d_j 
\end{equation}
where $i,j = 1,.., K$, i.e., total number of clustered classes, and $d_i$, $d_j$ indicate the first and second largest distances from certain centroids. Intuitively, lowest margin points imply a higher statistical uncertainty. Thus, the N margin values are sorted in ascending order and selected greedily up to exhausting a user-determined batch size $n$. Further, data points are added to the batch only if their cluster distribution mirrors the overall cluster distribution. This dynamic batch selection step accommodates for the second requirement, i.e., preserving diversity over the underlying data distribution. 

\subsection{Class Probability Propagation}
To integrate the knowledge on labeled classes produced on each Active Learning query, the annotated examples are used to train a supervised classifier. The classifier will then estimate probabilities for the remaining unlabeled points. The generated probability vectors are then concatenated to the original vector representation to update K-Means centroids accordingly. Further, higher weights are associated to the probability vectors, as opposed to the other features composing the input, to purposefully boost the incidence probabilities estimated from the human-annotated labeled subset.\\
Two classifiers are embedded in the discussed framework to test for their impact on the overall performance: Extreme Gradient Boosting (XGBoost) and Support Vector Machines (SVM). 
\begin{description}
\item[XGBoost]{Extreme Gradient Boosting \cite{friedman2001greedy} is a supervised model that generates predictions from input n-dimensional vectors, based on the observed classes. It summarizes judgments from an ensemble of classification and regression trees (CART) that carry decision values on the leaf nodes. Operationally, the predictions are summed across different trees, in the attempt to increase the accuracy of the classifier. Thus, while the mathematical modeling of XGBoost is mostly comparable to Random Forest models, the actual training routine implemented in the two cases is different. In fact, the so-called tree boosting routine optimizes performance (defined through a specific objective function) at each iteration, to refine the training of each subsequent decision tree.}
\item[SVM]{Given input n-dimensional vectors, an SVM classifier \cite{cortes1995support} estimates an hyperplane that can best group the data points into distinct classes. As a result, linear SVM does not ensure optimal separation for non-linear data. Typically, one adopted countermeasure is kernel-based learning, i.e., mapping the input vector features into a different mathematical space, to improve their separability \cite{scholkopf2002learning}. In the discussed framework, we integrated one commonly-used kernel setting based on the Radial Basis Function (RBF). This transformation function is derived from computing the radial Euclidean distance between points and centroids and normalizes the input vectors. }
\end{description}

\section{Experiments}
\subsection{Data Preparation}

\begin{description}
\item[Data collection.]{ The reference data set for these analyses is the result of an ongoing study of media behavior that captured device use across 101 participants located in the US, collectively providing for 5'532'739 smartphone screenshots. Full-resolution screenshots are taken every 5 seconds, bundled and encrypted for secure transmission to a Cloud-based centralized database. Images are organized in buckets (i.e., by subject ID) and associated with the related timestamps of capture. Screenshots are further processed and page layouts segmented to feed an Optical Character Recognition (OCR) module. Ultimately, one JSON document is produced for each input screenshot to store the extracted textual contents, when present.\\  
From the overall set, we selected a randomized sample, stratified across all subjects, by applying reservoir sampling with batch size 500. We chose reservoir sampling as randomization strategy (i) because each subject provided for a different amount of screenshots and (ii) to optimize selection across a significantly large collection. As a result, either 500 images were picked from each bucket (i.e., subject ID), when the considered set exceeded 500 units, or the whole set was taken otherwise. Once the randomized screenshot sample was identified, the related JSON documents enclosing auxiliary metadata and textual contents were retrieved as well.}
\item[Image Processing.]{ The selected screenshot sample is representative of different subjects, during consistent use captured at different time points. Therefore, we expect that even images that would be labeled by humans as belonging to the same cluster would enclose some degree of background and content heterogeneity. To normalize our image distribution and make it less sensitive to marginal noise, we first represented screenshots as 1280x720 grayscale-pixel vectors. \\
Vectors were further resized to 256x256 resolution and the embedded values ultimately standardized to a normal distribution. }
\item[Visual Feature Description.]{ To represent screenshots through a set of distinctive visual components, we adopted a state-of-the-art method for SIFT Object Recognition, i.e., Histogram of Gradients (HOG), as Visual Feature Descriptor. Thus, a grid of 8x8 pixel cells is built on top of the 256x256 images, to compute pixel-wise gradients and group them together based on 9 orientation bins. We set the sliding window size to blocks of 2x2 cells, to obtain a flattened representation of each screenshot as a one-dimensional vector consisting of 34'596 distinct features.}
\item[Dimensionality Reduction.]{ Vector representations obtained from the previous step were then reduced to a more compact form, to optimize computation across a large sample set. We initially applied Principal Component Analysis (PCA) over the first 1000 components to assess the optimal number of eigenvalues for the subsequent runs. To optimize the computational cost and leverage memory constraints for applying PCA to a high-dimensional matrix, we chose the randomized Singular-value Decomposition (SVD) option based on the work by Halko et al. \cite{halko2011finding}. We converged towards this specific number after inspecting the cumulative sum of variances explained by the top-n components. As a result, the selected threshold indicates the cutoff after which adding additional components does not lead to increased cumulative variance. The number of discovered principal components was then equal to 225 in the standalone image vectors case and equal to 897 in the joint image-text representation scenario.
}

\item[Text Processing.]{For each input screenshot, the textual content extracted through a combination of Image Processing and OCR is also retrieved, when available. First, we applied whitespace-based tokenization to the raw text and removed punctuation. Then, all the resulting tokens were reduced to lowercase. }
\item[Text Vector Representation.]{To obtain a vectorized representation of the text that was similarly dense, when compared to the aforementioned visual feature vectors, we utilized GloVe pre-trained 300-dimensional word embeddings derived from the Wikipedia 2014 and Gigaword 5 data sets and accounting for 6 Billion tokens in total. Word embeddings are combined through the following workflow: (i) for each token in the screenshot text, the corresponding semantic vector is retrieved, if the considered token is found in the reference vocabulary, (ii) a zero-valued vector is associated to each unknown token, (iii) the overall vector describing the full document is obtained from the weighted average of word embeddings identified in the first phase, using TF-IDF values as weights. Finally, (iv) zero-valued vectors are associated to all the screenshots that did not contain any text. }

\end{description}

\subsection{Parameter Sensitivity Analysis}
\begin{figure}[h]
\centering
\caption{\label{fig:elbow} SSE average values for $0 < K \leq 1000 $ over iterations of 10 units.}
\includegraphics[scale=0.6]{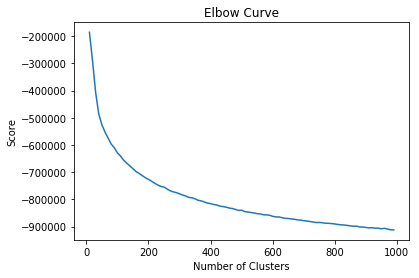}
\end{figure}

To aid the choice of K to apply K-Means, we run sensitivity analysis for this parameter with respect to the Sum of Squared Errors (SSE) curve.
For algorithms using Euclidean distance, accuracy can be derived from the sum of the squared errors (SSE), based, in this particular case, on the distance between data items and their centroids. Specifically, the SSE values are averaged across all clusters in the configuration.\\
We applied the elbow method for K selection on the SSE curve, after drawing the curve on a 0-1000 range in iterations of 10 units, as illustrated in Figure \ref{fig:elbow}. In brief, we selected the value of K corresponding to the 80\% break of the SSE curve, as the marginal gain after the identified point would not be sufficient to justify a higher number of clusters. In our case the selected K corresponds to 190 clusters.

\subsection{Experimental Settings}
In the discussed experiments, we exploited K-Means++ for centroid initialization and set K to 190, as explained in Section 4.2. In all the experimented setups, similarity of points is defined with respect to Euclidean distances. \\
Informative and Diverse Active Learning batch size was set to 200 over 10 iterations, leading to 2000 labels produced by the oracle, for each experimental pipeline.\\ 
Three different pipelines were run based on different combinations of input vector features and supervised classifiers used to estimate class probabilities for the unlabeled points on iterations. Specifically, (i) XGBoost and (ii) SVM classification was applied on the vectorized images, and, ultimately, (iii) the same workflow was tested over the joint image-text vector space as well.  

Screenshots selected by the Active Learning module were presented to two Graduate Students in Informatics and Industrial and Manufacturing Engineering with prior knowledge of the data characteristics and both attaining to a given taxonomy, consisting of 60 classes. Labels were formulated at the app-level of granularity (e.g., Facebook, Twitter, Reddit), or, when the content could be associated with any smartphone app, based on the type of action being represented (e.g., Navigation, Web search, Notifications). Sample screenshots by category are showcased in Figure \ref{fig:categories}.


\subsection{Evaluation}
Given the semi-supervised setting, internal cluster validity metrics were used to assess the impact of both the input representation (i.e., not including text vectors/including text vectors) and the embedded classifier (i.e., XGBoost, SVM) for probability propagation from the labeled subsample. Quality of the output configuration, across all three indices, is formulated by jointly measuring intra-cluster cohesion and inter-cluster separation.

\begin{description}

\item[Silhouette Index]{Introduced in \cite{rousseeuw1987silhouettes}, the silhouette index is used for graphical evaluation of the within-cluster cohesion, when leveraged with adequate separation across different clusters. For each data point i in the input set, let $a(i)$ be the average distance between i and all the remainder points lying in the same cluster. Further, one can derive $b(i)$ as the minimum average distance between i and all points in any other cluster. The Silhouette value for i is then defined as:

\begin{equation}
s(i) = 
\begin{cases}
      1 - a(i)/b(i), & \text{if}\ a(i)<b(i) \\
      0, & \text{if}\ a(i)=b(i) \\
      b(i)/a(i) - 1 , & \text{if}\ a(i)>b(i)
\end{cases}
\end{equation}

Thus, averaging $s(i)$ across all data points provides estimates about the overall quality of the clustered configuration.
}
\item[Dunn Index]{Another validity index is the Dunn Index \cite{dunn1974}. For each cluster of n-dimensional vectors $C_i$, where $k=1,...,K$, $\Delta_i$ indicates the maximum inter-vector distance:

\begin{equation}
\Delta_k = \max_{x,y \in C_k} d(x,y)
\end{equation}
}
Further, let $\delta(C_i, C_j)$ be the minimum distance between clusters $C_i$ and $C_j$ (formulated here as the distance between the two closest data points in the two clusters):

\begin{equation}
\delta(C_i,C_j) = \min_{x \in C_i, y \in C_j} d(x,y)
\end{equation}

The Dunn-Index mathematical formulation for K clusters can be then derived as:

\begin{equation}
DI_K =\frac{ \min_{1\leq i<j \leq K} \delta(C_i,C_j)}{ \max_{1 \leq k \leq K} \Delta_k} 
\end{equation}

\item[Davies-Bouldin Index]{Similarly to the Silhouette and Dunn indices, the Davies-Bouldin Index \cite{davies1979cluster} is another internal clustering evaluation metric. In the Euclidean distance scenario, let $C_i$ be a cluster of n-dimensional vectors where $i=1,...,N$, $S_i$ measures the \textit{scatter }within each cluster the Euclidean distance between vectors and their centroid. 
Further, for two clusters $C_i$ and $C_j$, let $M_{i,j}$ indicate the Euclidean distance between the two cluster centers as proxy of separation between $C_i$ and $C_j$. The Davies-Bouldin Index for K clusters is then derived, as a function of the ratio between the two said components:

\begin{equation}
DB_K = \frac{1}{K} \sum_{i=1}^{K}{\max{\frac{S_i + S_j}{M_{i,j}}}}
\end{equation}
}

\end{description}

By definition, the Silhouette and Dunn indices are maximized to improve the end cluster quality, whereas, conversely, we identified the local minimum for the Davis-Bouldin curve. 

\subsection{Experimental Results}
\begin{figure}[htbp]
 
    \begin{subfigure}[b]{0.5\textwidth}
      \caption{\label{fig:imgmatsvm} Visual feature vectors with SVM classification.}
       \includegraphics[width=\textwidth]{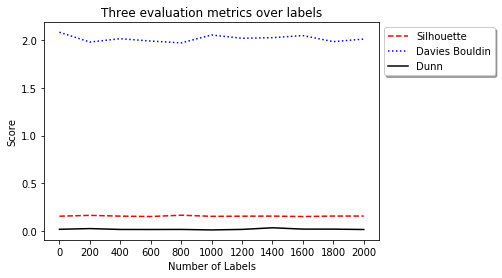}
    \end{subfigure}
    \begin{subfigure}[b]{0.5\textwidth}
       \caption{\label{fig:imgmatxgb} Multi-modal feature vectors with XGBoost classification. }
       \includegraphics[width=\textwidth]{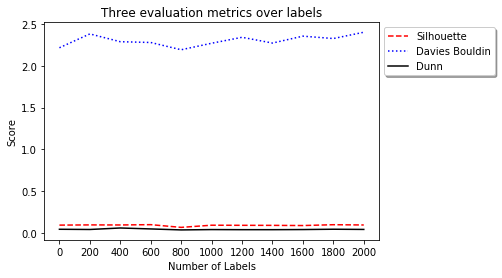}
    \end{subfigure}
    \begin{subfigure}[t]{\textwidth}
        \centering
       \caption{\label{fig:imgtextmatxgb} Visual feature vectors with XGBoost classification.}
       \includegraphics[width=0.5\textwidth]{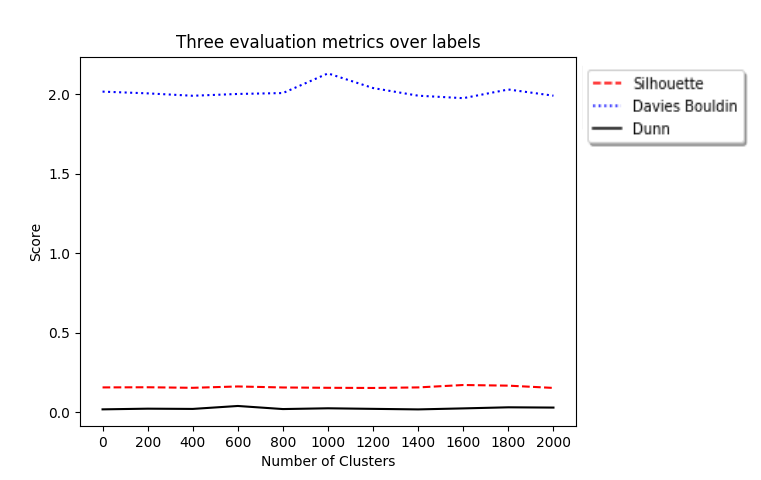}
    \end{subfigure}
    
    \caption{\label{fig:results} Cluster validity results, evaluated w.r.t. Silhouette, Davies-Bouldin and Dunn indices,over the numbered of labeled data points.}
\end{figure}

First, we applied the framework to the image vectorized representation (i.e., without including the text representation). Figure \ref{fig:imgmatsvm} shows the results obtained when integrating an SVM classifier with RBF kernel, over 10 total iterations and for incremental batches of 200 labels. After an initial marginal improvement from the starting scores, cluster validity keeps fluctuating in the same range of the score obtained with the first unsupervised configuration, before introducing any labeled examples. \\
We obtained comparable results across all evaluation metrics when repeating the experiment for the same input features, but exploiting a XGBoost-embedded module instead (Figure \ref{fig:imgmatxgb}). \\
Ultimately, the joint image-text embeddings were given as input to the XGBoost-based solution for semi-supervised clustering. As depicted by Figure \ref{fig:imgtextmatxgb}, introducing the textual content led to a performance decrease according to all the three evaluation metrics. This result indicates that although potentially helpful to disambiguate images enclosing the same set of application icons but presenting different backgrounds (e.g., in the "Home screen" case, Figure \ref{fig:home}), the current representation introduced additional entropy when clustering points in this alternative vector space. Said result could also be related to the specificity of the pre-trained corpora, when contrasted to the peculiar and ad hoc vocabularies found on smartphone screens.  

\section{Conclusion}
This paper presented exploratory analyses on clustering smartphone screenshots based on multimodal (visual and textual) features. Semi-supervised clustering was applied in combination with two alternative supervised classifiers, to propagate class probabilities from the labeled subset. Manually annotated labels were produced through active learning methods, i.e., after informative and diverse batch selection. \\
We found that performance, expressed in terms of cluster validity, was comparable with respect to the employed classifier and can, by all means, be further ameliorated. Further, the current representation of textual features, on average, led to a slight performance decrease. All cases, nonetheless, showed a quite persistent performance even after increasing the size of the labeled sample.\\
All the said evidences will inspire two major directions of future improvement: (i) revising the strategies and methods used to create \textit{ad hoc} image-text embeddings, (ii) refining the taxonomy used for annotating screenshots, mirroring the newly-formulated representation. Further, in the experimented setting, labels were simply provided by human annotators, based on the given taxonomy and on their sole judgement. However, refinements in representing and classifying the content could possibly lead to providing human judges with a set of top candidate label suggestions to choose from, to further leverage manual and automated categorization and in the attempt to minimize the individual bias. \\
The seminal experiments presented in this paper have confirmed the challenging nature of classification in the specific case of smartphone screenshots, justifying further investigation based from the lessons learned during this exploratory assessment.

\section{Acknowledgments}
We thank all the Screenomics Lab (\url{http://screenomics.stanford.edu/}) members for the useful discussions and acknowledge the data and computational
support provided for these experiments.

\bibliographystyle{unsrt}
\bibliography{references}

\end{document}